\documentclass[12pt]{article}

\usepackage{authblk}
\newcommand*\samethanks[1][\value{footnote}]{\footnotemark[#1]}

\newcommand{\my}[1]{{\color{black}{#1}}}

\usepackage[
backend=biber,
style=numeric-comp,
alldates=year,
url=false,
]{biblatex}
\usepackage{amsmath}
\usepackage{amssymb}
\usepackage{graphicx}
\usepackage{geometry}
\usepackage{color}
\def\Psc{\mathcal{P}}
\usepackage{setspace}

\usepackage{longtable}
\title{\vspace{-2em} TRACER: Transfer Learning based Real-time Adaptation for Clinical Evolving Risk}
\date{} 
\author[1,2]{Mengying Yan, PhD}
\author[1]{Ziye Tian, BS}
\author[3,4]{Siqi Li, PhD}
\author[1,3,4,5,6]{Nan Liu, PhD}
\author[1,9]{Benjamin A. Goldstein, PhD \thanks{Co-last authors contributed equally to this work. Correspondence to: ben.goldstein@duke.edu (B. A. Goldstein),  moleiliu95@gmail.com (M. Liu), and
Chuan.Hong@duke.edu (C. Hong).}}
\author[7,8]{Molei Liu, PhD \samethanks[1]}
\author[1,9]{Chuan Hong, PhD \samethanks[1]}

\affil[1]{Department of Biostatistics and Bioinformatics, Duke University School of Medicine, Durham, NC, USA}

\affil[2]{Duke AI Health, Duke University School of Medicine, Durham, NC, USA}

\affil[3]{Centre for Quantitative Medicine, Duke-NUS Medical School, Singapore, Singapore}
\affil[4]{Duke-NUS AI + Medical Sciences Initiative, Duke-NUS Medical School, Singapore, Singapore}

\affil[5]{Pre-hospital \& Emergency Research Centre, Health Services and Systems Research, Duke-NUS Medical School, Singapore, Singapore}

\affil[6]{NUS Artificial Intelligence Institute, National University of Singapore, Singapore, Singapore}

\affil[7]{Department of Biostatistics, Peking University Health Science Center, Beijing, China}

\affil[8]{Beijing International Center for Mathematical Research, Peking University, Beijing, China}

\affil[9]{Duke Clinical Research Institute, Durham, NC, USA}

\begin{document}
\maketitle






\newpage
\section*{ABSTRACT}

\textbf{Objectives:}
Electronic health records-based clinical decision support tools are useful for medical decision-making. However, models trained on retrospective cohorts may suffer from performance drift when applied to current systems due to covariate shift and model shift over time. Especially, such temporal change may result from a leading event such as a systematic operation update or emerging diseases (e.g., COVID-19). A model updating approach is needed to account for the transition which occurring at different times for each individual.

\textbf{Materials and Methods:}
We propose a novel framework TRACER (Transfer Learning based Real-time Adaptation for Clinical Evolving Risk) that identifies individual-level temporal transition and leverages state-of-the-art transfer learning method to account for temporal transition without retraining the full models. We evaluate TRACER through simulation studies comparing its performance against static models. As an application, we consider model transition before and after COVID-19 for predicting hospital admission after emergency department (ED) visits at Duke University Health System. 

\textbf{Results:}
Simulation studies demonstrated that TRACER achieved better predictive performance compared to static models trained only on historical or current data. In the real-world application, TRACER improved AUC, Brier score and $R^2$ over baseline models, effectively mitigating performance drift across the COVID-19 transition.

\textbf{Discussion:}
By incorporating individual-level transition detection with transfer learning, TRACER provides a scalable and flexible solution to maintain predictive performance under evolving clinical conditions.

\textbf{Conclusion:}
TRACER improves model adaptability to temporal changes and offers a practical strategy to enhance the reliability of EHR-based decision support tools in dynamic healthcare environments.

\textbf{Keywords:}  Transfer learning; Clinical prediction model; Performance drift; Electronic Health Records; Clinical evolving risk

\newpage
\section{INTRODUCTION}


Clinical decision support tools (CDSTs) play an important role in modern healthcare by leveraging electronic health records (EHR) data to guide clinical decision-making processes \cite{alexiuk_clinical_2024,goldstein_opportunities_2017}. These tools are used to predict patient outcomes, identify high-risk individuals, and customize treatment plans. Numerous healthcare institutions have already integrated EHR-based CDSTs to enhance quality of care, reduce costs, and streamline clinical workflows \cite{parikh_integrating_2016,sutton_overview_2020,gracia_martinez_implementing_2025}. However, clinical prediction models can suffer from performance drift when there are changing practice pattern, patient populations and risk conditions.
\cite{davis_calibration_2017,finlayson_clinician_2021,moons_risk_2012,subbaswamy_development_2020}. This performance drift can reduce model performance and affect clinical decision support if not adequately addressed.

Such temporal shift includes covariate shift and model shift. Covariate shift occurs when the distribution of predictors changes, for example, when patient demographics shift. Model shift arises when the relationship between predictors and outcomes evolves, which is sometimes triggered by wide-ranging events such as system upgrades, policy changes, or emerging diseases like COVID-19 and flu. Although retrospective models may work well initially, their predictive power may decline once the underlying healthcare environment changes.
\my{Temporal shifts in clinical data rarely shift
the whole population at once. Instead, only a subset of patients is affected by the
precipitating event, while the remaining patients continue to resemble the historical cohort. In this work, we focus on such case-mix shifts, where changes occur in only a portion of the patient population. For example, changes may arise from the staggered rollout of new clinical protocols, heterogeneous adoption of telehealth, or patient-level variation in exposure to new clinical workflows.}

A variety of strategies have been proposed to address temporal changes in predictive modeling. Traditional approaches include updating existing models through recalibration, coefficient revision, or shrinkage techniques to adjust for changes in outcome prevalence or covariate distributions \cite{moons_risk_2012, steyerberg_validation_2004, davis_nonparametric_2019}. Bayesian updating methods \cite{xu_using_2014} incorporate prior knowledge from previously trained models into posterior distributions, allowing new data to refine the model without discarding historical information. Tree-based ensemble methods, such as those using staged gradient-boosted models and ensemble stacking, combine insights from both old and new data to improve generalizability across time. Continual learning approaches \cite{amrollahi_leveraging_2022} incrementally update models as new data arrives, preserving prior knowledge while adapting to recent patterns. Meta-learning frameworks, such as L2E, offer another line of work for handling dynamic data streams by learning to adapt models across evolving domains \cite{wu_unified_2022}.

We consider transfer learning methods \cite{weiss_survey_2016, li_bridging_2024} which leverage knowledge from a source domain to improve model performance in a target domain. These methods offer a promising approach to adapt existing models to new environments through targeted updates rather than complete retraining. One state-of-the-art method is Trans-Lasso \cite{li_transfer_2022}, which addresses distributional shifts by selectively penalizing model coefficients and borrowing strength from previously learned features. In addition, it is a regression-based method that provides good interpretability.

\begin{figure}[h]
    \centering
    \includegraphics[scale=0.8]{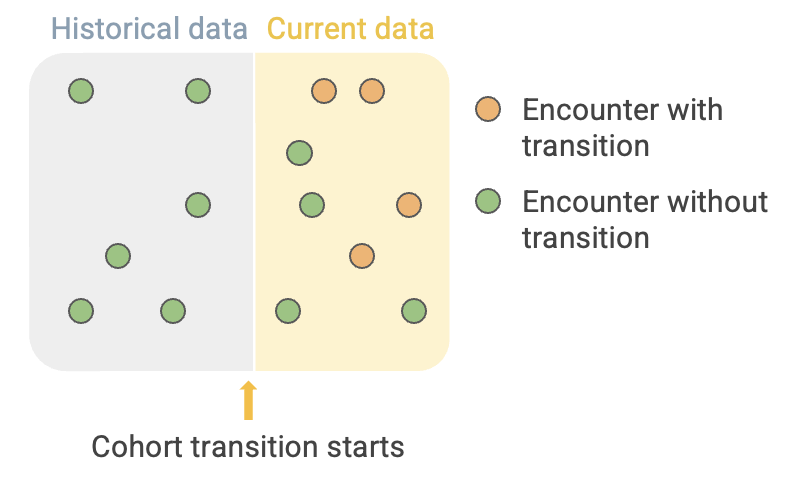}
    \caption{Mixed data pattern caused by temporal shift due to COVID-19 pandemic. We only know when the pandemic (transition) starts, but don't know the transition status of each encounter.}\label{fig:trans}
\end{figure}

Traditional transfer-learning approaches are typically designed for settings in which the source and target populations are entirely distinct, such as transferring a model from one hospital to another. 
\my{In contrast, we consider a setting in which only a subset of patients transitions to a new data distribution.}
\my{As illustrated in Figure \ref{fig:trans}, after such transition, the current data becomes a mixture of encounters from post-transition distribution and encounters from pre-transition (historical) distribution. Encounters from the pre-transition distribution remain similar to the historical cohort, whereas those from the post-transition distribution differ substantially. Therefore, the challenge is not only the temporal shift, but also identifying whether a given encounter is from the post- or pre-transition distribution. }
An additional challenge of adapting the model to the evolving target distribution is that we often only limited post-transition data, as was the case in the early stages of the COVID-19 pandemic.

To address the challenges of
(1) mixture of pre-transition and post-transition cohorts in the current data \my{due to heterogeneous timing of changes across subgroups}, and
(2) not having sufficient transitioned samples to train a full model, we propose a novel framework TRACER (Transfer Learning based Real-time Adaptation for Clinical Evolving Risk). TRACER uses an expectation-maximization (EM) algorithm to detect individual-level temporal transition status and incorporates Trans-Lasso to adapt to evolving risk without retraining the full model. As a practical demonstration, we evaluate TRACER on the task of predicting hospital admission after emergency department (ED) visits, using data from before and after COVID-19 as an illustrative temporal shift. We also conduct a series of simulations to evaluate the proposed method. Our findings demonstrate that TRACER maintains robust predictive performance under evolving clinical conditions, offering a practical solution to temporal drift in EHR-based clinical decision support tools.

\section{MATERIALS AND METHODS}

We first introduce the proposed framework TRACER in Section \ref{TRACER}. Then we will describe the simulation setting and real data application in Sections \ref{sim} and \ref{EHR}.


\subsection{TRACER}
\label{TRACER}
\subsubsection{Model setup}
Let $(Y_i,X_i)$ for  $i=1,2,\ldots,N$ denote the observed data for all encounters. 
We assume that an event can cause a potential shift (e.g., COVID-19 pandemic) starts at a certain time point. We are interested in having a model that adapt to the potential shift for the current cohort after that time point. We suppose that the historical data follows a historical source model, and any encounters in the current data can transition to a new model. Therefore, the observed current data include a mixture of encounters \my{from pre-transition distribution and encounters from post-transition distribution.}
The objective is to improve the model prediction performance in the mixed current data without retrain the historical model. 

Since we are interested in having the prediction in a timely manner, i.e. right after the transition happened, the sample size of the current data may be very small. We want a method that works well given relatively small sample size of current data. 

Let $S_i$ be the indicator of the \my{latent membership of pre- or post- transition distribution for encounter $i$.} 
This transition is characterized by the change of its data distribution $Y_{i},X_{i}$. 

Outcome model is a generalized linear model:

\begin{align}
P(y_i=1 | X_i, S_i=s)= g^{-1}(X_i\eta_s),
\end{align}


where $g$ is the link function, and $\eta_0$ and $\eta_1$ are the parameters for outcome model before and after transition, respectively. In a transfer learning regime, we assume that the gap between the two latent models $\eta^1-\eta^0$, is close to $0$. To leverage this assumption, we could re-formulate the parameters as:
\[
\quad\eta_0,~\delta_{Y|X}=\eta_1-\eta_0.
\]
Then inspired by Trans-Lasso \cite{li_transfer_2022}, for high-dimensional features, we impose penalties on $\eta_0$, and $\delta_{Y|X}(t)$  by introducing $$\lambda_1\Psc_1(\eta_0)+\lambda_2\Psc_2(\delta_{Y|X}).$$ We could consider the sparse $\ell_1$ penalty on $\Psc_1(\eta_0)$ and $\Psc_2(\delta_{Y|X})$.

Meanwhile, we assume $P(S_{i}=1)=\pi(W_i)$ is determined by some covariates $W_i$ (such as age, gender, COVID-19 diagnosis). We can specify the prior probability model for \my{latent membership change}
$\pi(W_i)$ as some generalized linear model with parameter $\gamma$:
\begin{align}
    \pi(W_i)=g^{-1}(W_i\gamma),
\end{align}
where $g$ is the link function and $\pi(W_i)=0$ if $t_i<T$.

For covariates, we suppose that there is a subset of covariates $A$ such that the distribution of which may differ from each other before and after transition. For example, we know that there will not be transition on features like age and gender for each subject. Suppose that density  $f(A_i|W_i,S_i=0)$ and $f(A_i|W_i,S_i=1)$ can be parameterized with $\beta_0,\beta_1$.

\begin{figure}[h]
    \centering
    \includegraphics[scale=0.5]{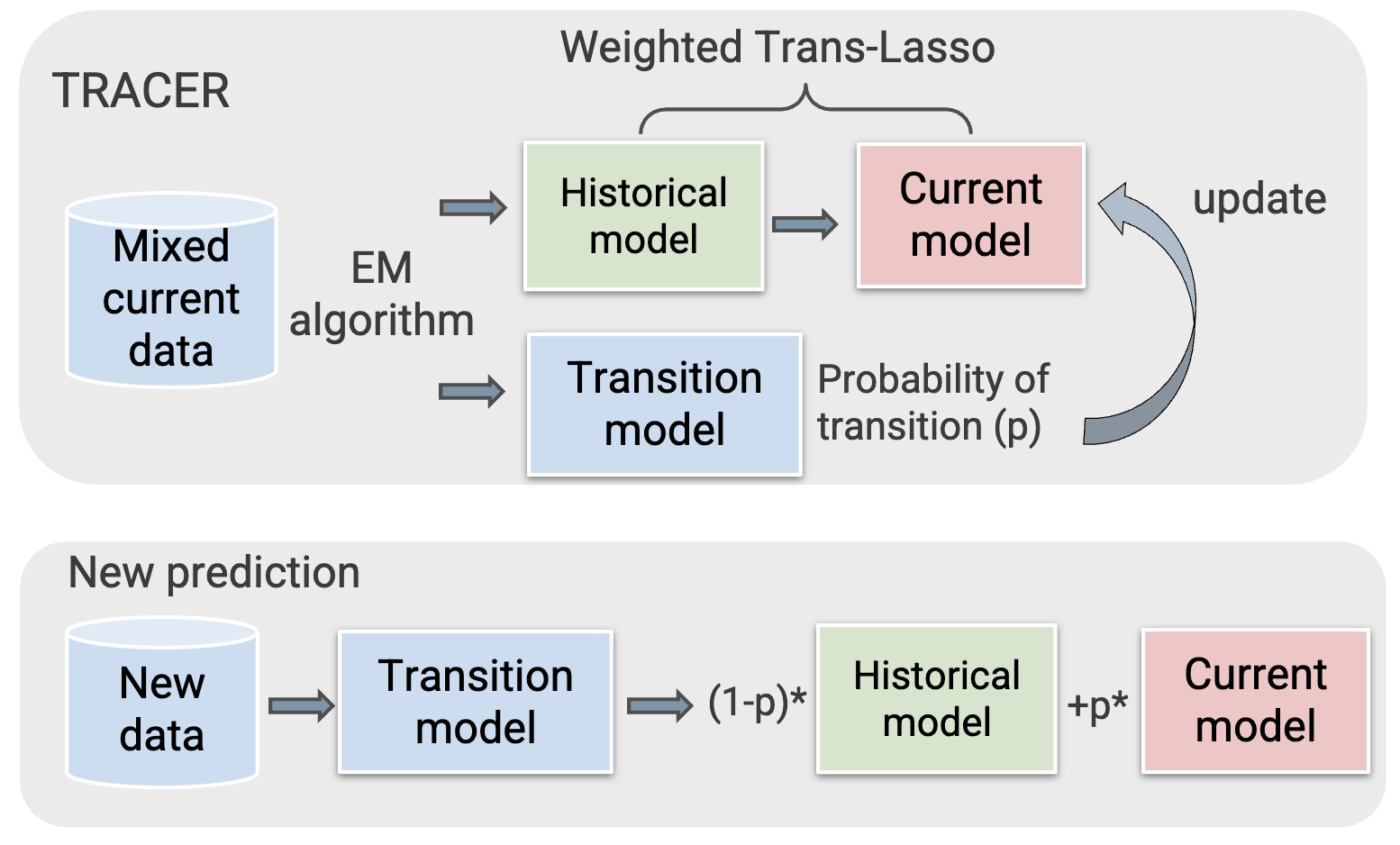}
    \caption{Proposed TRACER workflow. TRACER uses an EM algorithm to detect individual-level transitions and incorporates Trans-Lasso to adapt to evolving risk without retraining the full model}\label{method}
\end{figure}

The challenge is that $S_i$ is a latent indicator that is unknown in the observed data. As shown in Figure \ref{method}, we proposed to use EM algorithm to estimate the parameters for transition and then use the \my{probability of latent transition membership} 
as weight to estimate and update the shifted model simultaneously. In the EM algorithm, E step is calculated based on posterior probability $P(S_i|W_i, A_i,Y_i)$ and M step updates the parameters. In the end, we have estimated parameters $\hat\gamma, \hat\beta_0,\hat\beta_1,\hat\Sigma_0,\hat\Sigma_1,\hat\eta_0, \text{ and }\hat\eta_1$.  The details about the EM algorithm and its assumptions are shown in Appendix~\ref{sec:ass} and Appendix~\ref{sec:EM} . 

\subsubsection{Prediction}
For a new, coming record $\tilde X_{i}$ after $T$, the predicted outcome probability is
\begin{align*}
&\widehat{P}_{i\text{TRACER}}=\widehat{P}(S_i=1|W_i,A_i)\tilde X_{i}^T\widehat\eta_1+(1-\widehat{P}(S_i=1|W_i,A_i))\tilde X_{i}^T\widehat\eta_0,
\end{align*}

We also propose to use a small current data set as a validation set to form a weighted combined model when applying the proposed approach to real world data. We first obtain prediction for validation $X_{val}$ $\widehat{P}_{\text{TRACER}}(X_{val})$, $\widehat{P}_{\text{hist}}(X_{val})$, $\widehat{P}_{\text{current}}(X_{val})$ using TRACER, model trained only on historical data, and model trained only on current data, respectively. Then, regress $Y_{val}$ on $\widehat{P}_{\text{TRACER}}(X_{val})$, $\widehat{P}_{\text{hist}}(X_{val})$, $\widehat{P}_{\text{current}}(X_{val})$, and use the estimated coefficients $w_{\text{TRACER}}$,$w_{\text{hist}}$ and $w_{\text{current}}$ as weights to weight the final prediction.
$$\widehat{P}(\tilde X_{i})=w_{\text{TRACER}}\widehat{P}_{\text{TRACER}}(\tilde X_{i})+w_{\text{hist}}\widehat{P}_{\text{hist}}(\tilde X_{i})+w_{\text{current}}\widehat{P}_{\text{current}}(\tilde X_{i}).$$

\subsection{Simulation Setting}
\label{sim}
We evaluate the performance of TRACER using simulations. First, we generate covariates $W$ from multinormal distribution with mean 0.1 and identity matrix as covariate matrix. Next, the true 
\my{transition membership} $S$ is generated from binomial distribution
\begin{align}
   \text{Binomial}(\frac{\exp(\gamma^TW_i)}{1+\exp(\gamma^TW_i)}),
\end{align}
where the intercept is set to control ratio of pre-transition and post-transition data. We use three settings with intercept = (-3,-4,-5). Then, time-varying covariates $A$ are generated by Gaussian mixture: 
$$A_i|W_i, S_i=0 \sim N(\mu_0, \Sigma_0)=N(\beta_0^T W_i, \Sigma_0)$$ and $$A_i|W_i,S_i=1 \sim N(\mu_1, \Sigma_1)=N(\beta_1^T W_i, \Sigma_1).$$ For $S=0$ coefficients $\beta_0$, intercept is set as 0.1 and the rest coefficients as 0.1. For $S=1$ coefficients $\beta_1$, intercept is set as 0.1 and the rest as 0.3. We also add 50 or 100 zero coefficients to add sparsity.  

Finally, outcome $Y$ is generated using outcome models
$$\text{Binomial}(\frac{\exp(\eta_0^{T}X_i)}{1+\exp(\eta_0^{T}X_i^)})$$ for pre-transition and $$\text{Binomial}(\frac{\exp(\eta_1^{T}X_i)}{1+\exp(\eta_1^{T}X_i^)})$$ for post-transition, where $\delta=\eta_1-\eta_0$ is set to control model shift between pre- and post-transition models. We $\delta=(0.2,0.4,0.6)$
We compare area under the receiver operating curve (AUC) and mean squared error (MSE) for TRACER, model based on historical source data and model based on mixed current data.

\subsection{Real Data Application}
\label{EHR}
We evaluate the performance of our proposed approach using Electronic Health Records (EHRs) data from the Duke University Health System (DUHS), accessed through our research ready datamart \cite{hurst_development_nodate}. We are interested in predicting whether an emergency department (ED) visit will result in hospital admission (binary). We included 107 predictors in the model, including demographics information: age and sex; vital signs: pulse (beats/min), systolic blood pressure (SBP; mm Hg), diastolic blood pressure (DBP; mm Hg), oxygen saturation (SpO$_2$; \%), temperature ($^{\circ}$F), respiration (times/min) and acuity level; and comorbidities that defined based on the ICD codes and mapped to Phecode\cite{denny_systematic_2013}, a full list of variable used is included in the Appendix \ref{predictors}. Table~\ref{table1} shows the description of the study cohorts with standardized mean differences (SMD). Most of the SMDs exceed 0.1, indicating meaningful differences between the historical and current cohorts. This suggests that directly applying the historical model to the current cohort may lead to performance drift.

\begin{table}[htbp]
    \centering
    \begin{tabular}{lccc}
        \hline
        & \textbf{Historical} & \textbf{Current} & \textbf{SMD} \\
        & \textbf{(N=367,201)} & \textbf{(N=7,293)}  & \\
        \hline
        \textbf{Outcome} & 78,697 (21.43\%) & 2,962 (40.61\%) & 0.420 \\
        \hline
        \textbf{Age}, mean (sd) & 43.11 (23.03) & 38.03 (26.70) & 0.197 \\
        \hline
        \textbf{Sex} & & & \\
        \hspace{1em} Female & 211,273 (57.54\%) & 3,652 (50.08\%) & 0.149 \\
        \hspace{1em} Male & 155,928 (42.46\%) & 3,641 (49.92\%) & 0.149 \\
        \textbf{Systolic blood pressure}, mean (sd) & 134.03 (22.56) & 129.16 (20.29) & 0.229 \\
        \hline
        \textbf{Diastolic blood pressure}, mean (sd) & 79.48 (14.05) & 77.40 (13.95) & 0.148 \\
        \hline
        \textbf{Pulse}, mean (sd) & 89.53 (21.16) & 114.05 (26.77) & 0.987 \\
        \hline
        \textbf{Oxygen saturation}, mean (sd) & 98.11 (2.56) & 97.36 (3.36) & 0.240 \\
        \hline
        \textbf{Temperature}, mean (sd) & 98.25 (1.92) & 101.32 (1.24) & 1.827 \\
        \hline
        \textbf{Respiration}, mean (sd) & 18.80 (4.53) & 21.50 (6.83) & 0.474 \\
        \hline
        \textbf{Acuity level} & & & \\
        \hspace{1em} 1 & 3,067 (0.84\%) & 41 (0.56\%) & 0.031 \\
        \hspace{1em} 2 & 84,550 (23.03\%) & 2,589 (35.50\%) & 0.275 \\
        \hspace{1em} 3 & 183,937 (50.09\%) & 3,200 (43.87\%) & 0.125 \\
        \hspace{1em} 4 & 83,511 (22.74\%) & 1,338 (18.35\%) & 0.114 \\
        \hspace{1em} 5 & 13,136 (3.31\%) & 125 (1.71\%) & 0.103 \\
        \hline
        \textbf{Local tumor} & 3,773 (1.03\%) & 185 (2.54\%) & 0.116 \\
        \hline
        \textbf{Metastatic tumor} & 7,612 (2.07\%) & 327 (4.48\%) & 0.133 \\
        \hline
        \textbf{Diabetes with complication} & 26,061 (7.10\%) & 701 (9.61\%) & 0.087 \\
        \hline
        \textbf{Diabetes without complication} & 50,469 (13.74\%) & 1,204 (16.51\%) & 0.076 \\
        \hline
        \textbf{Renal disease} & 61,741 (16.81\%) & 1,740 (23.86\%) & 0.174 \\
        \hline
    \end{tabular}
    \caption{Description of the study cohorts with standardized mean differences (SMD).}
    \label{table1}
\end{table}

We use all the ED encounters in 2018 and 2019 as historical source population. We define the current population as 2020 ED encounters with a recorded temperature above $100^{\circ}$F, as fever was a hallmark COVID-19 symptom and a primary trigger for modified triage and infection control procedures during early pandemic months. Focusing on febrile encounters isolates the subset most impacted by altered protocols, resource constraints, and shifts in care-seeking behavior, thereby amplifying the relevance and detectability of the pandemic-related distributional shift. 
\my{These changes did not occur simultaneously across the entire patient population but instead emerged within a subset of encounters.} 
To simulate early-pandemic conditions with limited data availability, we use monthly samples from March to September 2020 as the currently available data and reserve October to December 2020 as the test set. Sample sizes for March to September in 2020 are 694, 375, 458, 590, 591, 550 and 457. Test set has sample size 1422. This setup mimics the scenario shortly after the onset of COVID-19, when only a small amount of data is available, and rapid model adaptation is needed.
We apply the proposed TRACER methods to address the distribution shift, and compare their performance against three baselines: (1) a model trained solely on historical data (historical model), (2) a model trained solely on current data that is a mixtrue of pre-transition and post-transition models (current only model), and (3) a model using TransLasso applied directly to the current data without accounting for its mixed distribution. We evaluate all methods using AUROC, Brier score and $R^2$.

\section{RESULTS}
\subsection{Simulation Results}
\begin{figure}[htb]
    \centering
    \includegraphics[scale=0.6]{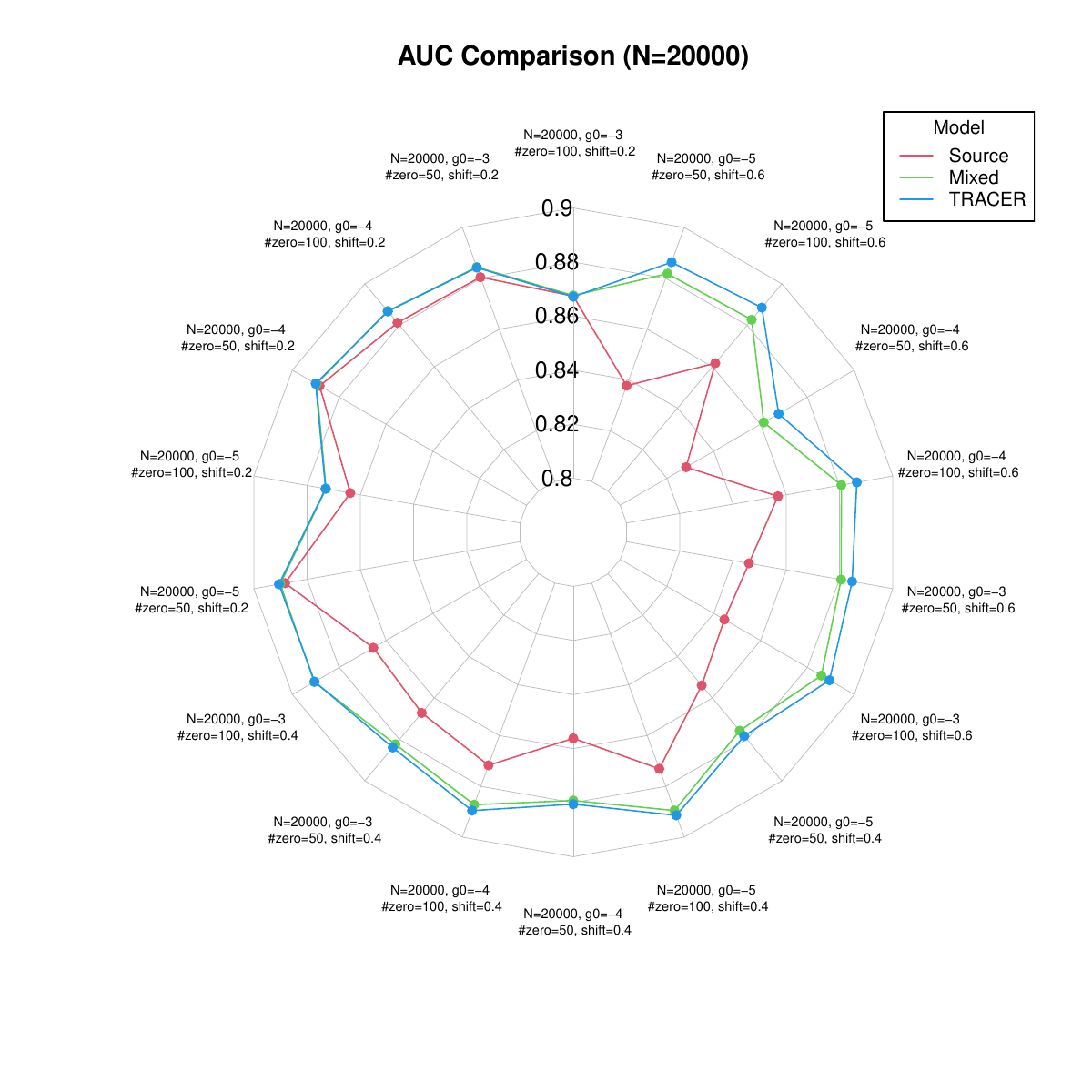}
    \caption{AUC comparisons across different models in the simulation study. Source/historical model: a model trained solely on historical data. Mixed/current model: a model trained solely on current pooled pre-transition and post-transition data. }\label{sim_auc}
\end{figure}

\begin{figure}[htb]
    \centering
    \includegraphics[scale=0.6]{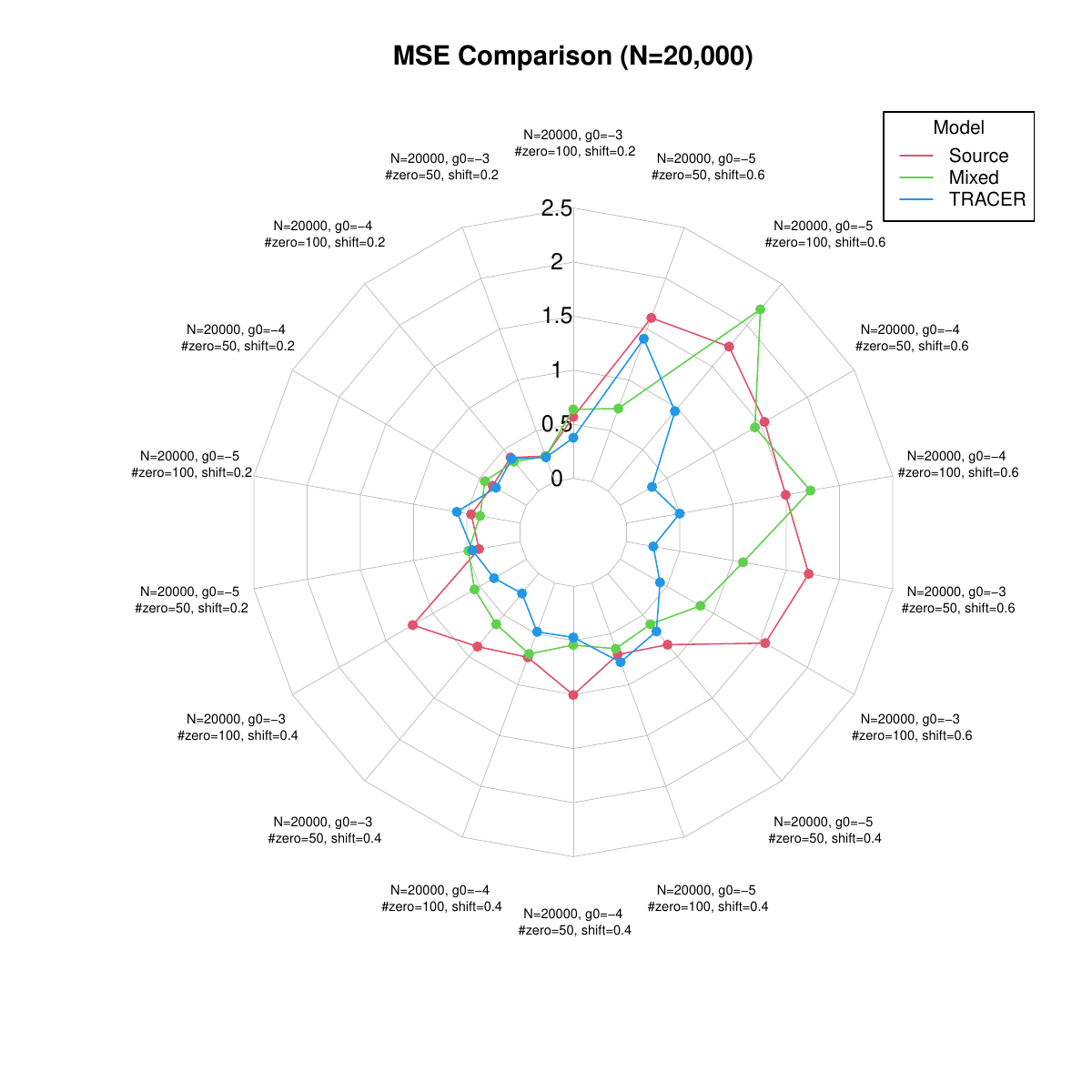}
    \caption{MSE comparisons across different models in the simulation study. /historical model: a model trained solely on historical data. Mixed/current model: a model trained solely on current pooled pre-transition and post-transition data.}\label{sim_mse}
\end{figure}

We evaluate the performance of TRACER using a series of simulation experiments that simulate covariate and model shifts over time. The simulation design allows variation in the proportion of pre-transition and post-transition data (via the transition model intercept $g_0$), the sparsity of covariates (via the number of zero coefficients), and the severity of model shift (via a shift parameter $\delta$). We compare three modeling approaches: a model trained solely on historical data (Source/Historical model), a model trained on current pooled pre-transition and post-transition data (Mixed/Current model), and our proposed method, TRACER, which adapts to temporal transitions at the individual level without full retraining.

The AUC comparison (Figure~\ref{sim_auc}) demonstrates that TRACER consistently outperforms the historical source model and generally matches or exceeds the performance of the current data mixed model. This pattern holds across all settings, including those with severe model shifts ($\delta = 0.6$) and varying degrees of sparsity (the number of zero coefficients = 50 or 100). While the historical model's performance deteriorates under larger shifts and more complex scenarios, TRACER remains robust, highlighting its adaptability and transportability. Notably, in high-shift or high-sparsity settings, the current only model also begins to lose effectiveness, whereas TRACER continues to maintain a high level of AUC performance, showing its strength in addressing individual-level transitions.

Complementing these findings, the MSE comparison (Figure~\ref{sim_mse}) reveals that TRACER achieves the lowest mean squared error across most scenarios. This is particularly evident in simulations with stronger model shift and greater sparsity, where both the historical and current only models exhibit higher error rates. In these settings, TRACER maintains a clear performance advantage, demonstrating its ability to reduce prediction error even when the data distribution has shifted substantially. The historical model performs poorly across most conditions, especially when the mismatch between training and current distributions is large. The current model, while often outperforming historical, is still less effective than TRACER, likely due to its inability to account for nuanced individual-level shifts.

Together, the AUC and MSE results confirm that TRACER offers a robust and efficient solution for adapting predictive models to temporal changes without the need for full retraining. It is particularly effective in scenarios characterized by distributional shifts and sparse high-dimensional covariates which are settings commonly encountered in real-world healthcare data. These findings underscore the potential of TRACER to support reliable clinical decision-making over time, maintaining model performance as populations and care patterns evolve.

\subsection{Real Data Application Results}
\begin{figure}[htb]
    \centering
    \includegraphics[scale=0.55]{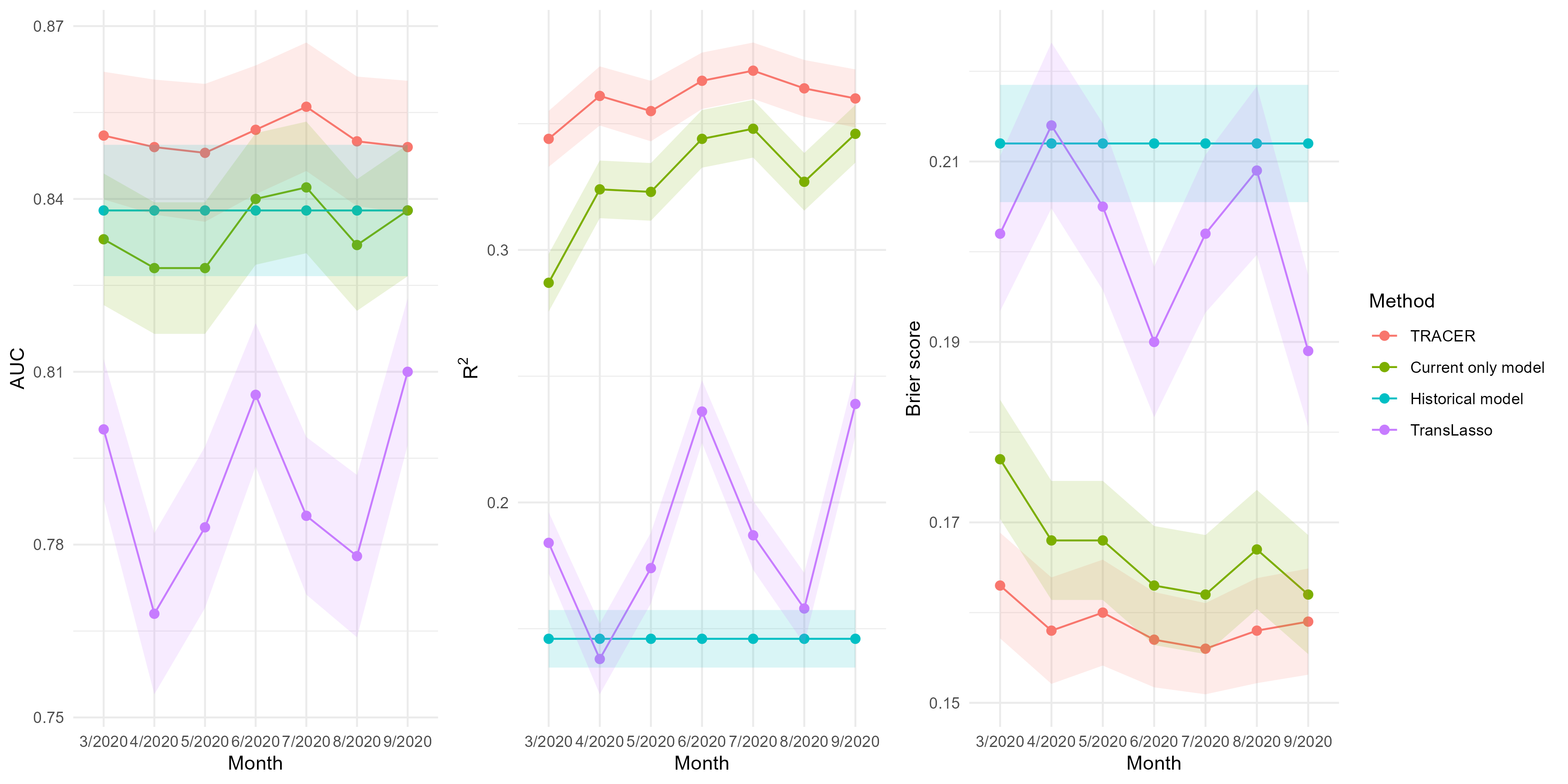}
    \caption{Comparison of predictive performance by month across different models in the real data application. Monthly performance metrics (AUC, Brier score, and R²) are shown for each model. Shaded bands represent 95\% bootstrap confidence intervals.}\label{ehr_res}
\end{figure}

Figure \ref{ehr_res} presents performance of four models (Historical model, Current only model, TransLasso, and TRACER) evaluated on EHR data from DUHS, to predict hospital admission following ED visits. 

The \textbf{Historical model}, trained on pre-COVID (2018-2019) data, achieved an AUC of 0.838, $R^2$ of 0.146 and Brier score of 0.212 when tested on the current 2020 cohort. The \textbf{Current only model}, trained only on one month data during COVID, yielded slightly improved $R^2$ (0.287–0.348) and Brier scores (0.162–0.177) while maintaining similar AUCs (0.828–0.842).
These differences between historical and current only models illustrate the impact of temporal distributional shifts on model. Relying solely on outdated data degrades calibration and explanatory power. However, exclusive dependence on limited contemporary data constrains generalizability due to small sample sizes.

The \textbf{TransLasso} model treats historical as the source and current data as the target but does not account for the mixture nature of the current cohort. It improved over the Historical model in Brier score and $R^2$, and outperformed the Current-only model in AUC. However, it did not achieve the overall best performance.

\textbf{TRACER}, which integrates predictions from both post-transition and the historical models, consistently outperforms all other methods. Across all months, TRACER achieved the highest AUCs (0.848–0.856), the highest $R^2$ values (0.344–0.371), and the lowest Brier scores (0.156–0.163). These results demonstrate the benefit of strategically combining historical and emerging data representations to leverage complementary information and mitigate the weaknesses of both single-source and small-sample models.



\begin{figure}[htb]
    \centering
    \includegraphics[scale=0.8]{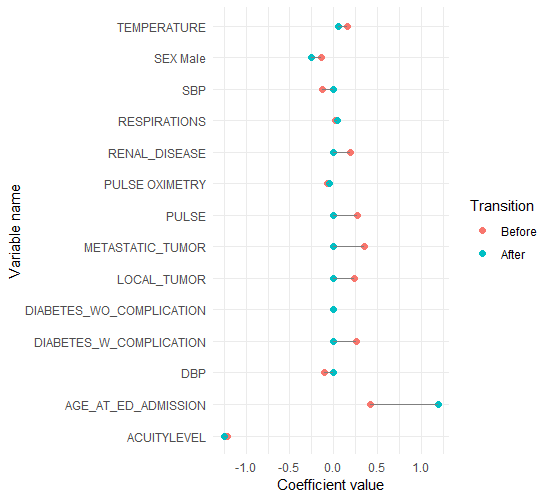}
    \caption{Comparison of coefficients before and after transition in TRACER. Coefficient estimates from the historical source model and post-transition model used in TRACER are shown. Each point represents the effect size of a predictor. Differences highlight changes in predictor importance across the transition.}\label{coef}
\end{figure}

Figure \ref{coef} displays selected coefficients from TRACER’s pre-transition and post-transition models, with the full set for all 107 predictors shown in Appendix \ref{coef_list}. Notably, several coefficients shift meaningfully between stages. For example, the post-transition model assigns a larger positive weight to age, suggesting that older patients were more likely to be admitted to the hospital after ED visits during COVID.

\section{DISCUSSION}
This study highlights the importance of addressing temporal shifts in clinical prediction models. We addressed the challenges of \my{a change of underlying data distribution in a subset of population, and not having sufficient transitioned samples to train a full new model.}
We introduce TRACER, a novel framework that combines individual-level 
\my{latent transition membership} identification with a weighted transfer learning approach. 
\my{TRACER models a transition process where the underlying data distribution for a subset of the population changes and it incorporates that change into an updating model without full model retrain.}
Simulation results demonstrated that the proposed method improves AUC and MSE compared to static models. In the real world data application, our method outperformed the baselines in AUC, $R^2$ and Brier score. It showed improvements in transportable performance under evolving healthcare conditions caused by events such as COVID-19 pandemic. In addition, TRACER maintained strong performance even in small-sample settings, which are common in abrupt healthcare shifts such as the beginning of a pandemic. 

TRACER does not simply assume that data shift occurs at the same time, which is often the case, for example, not all patients had COVID at the same time. \my{Instead we assume at any given time, only a subset of the population transitions to a new distribution, resulting in a mixture of encounters that remain similar to the historical data and encounters that deviate from it. }
This allows CPMs to adapt to different patient trajectories or phased implementations of new protocols. Moreover, TRACER leverages transfer learning and regression-based models, which provide both adaptability and interpretability. Trans-Lasso offers an interpretable and flexible alternative to more black-boxed methods. Such interpretability is particularly important in clinical settings, where model transparency can enhance trust and facilitate integration into decision-making workflows. 

Our framework integrates the transfer learning step into the EM algorithm to estimate my{latent transition membership} 
and update the outcome model simultaneously. In contrast to traditional strategies that rely on retraining, our method offers a computationally efficient solution that supports continuous learning. As new data becomes available, the framework can be used to update in real-time, preserving historical knowledge while adapting to ongoing changes in the healthcare system.

This additionally highlights the potential for proactive monitoring and maintenance of CDSTs within clinical settings. Static models often degrade over time due to covariate and model shifts, and TRACER offers a principled way to adjust model parameters while maintaining stable performance. \my{In practice, a key consideration is when to update the model and how to define the ``current" data used for adaptation. We suggest two practical strategies. First, a clinically driven cutoff can be specified based on domain knowledge, such as documented policy changes, protocol rollouts, or system-level interventions. Second, a performance driven cutoff can be employed, in which model updates are triggered by observed degradation in predictive performance, such as declines in discrimination (e.g., AUC) or calibration. }

While our work estimates the probability \my{of latent transition membership }
for each encounter, future research could extend this framework to accommodate multiple transitions in a longitudinal setting. For example, patients undergoing repeated encounters across different phases of care may require models that adapt to evolving risk over time. 
\my{Compared with fully dynamic updating approaches that require continuous retraining, TRACER offers a lightweight alternative that may be more practical for real-world deployment. A comparison to dynamic updating can be one of the future work.} 
\my{TRACER is primarily designed to address shift arising from changes in case mix, where only a subset of the population transitions to a new data-generating distribution. While the framework can partially accommodate association shifts and event-rate shifts through recalibration, it is not explicitly designed to handle label shift or concept drift, where the underlying outcome definitions or predictor–outcome relationships fundamentally change. Extending TRACER to more fully address these forms of distributional change represents an important direction for future work.}

\section{CONCLUSION}
In conclusion, TRACER provides a practical way to adapt CPMs to dynamic healthcare environments. By addressing temporal shifts with a state-of-the-art transfer learning method, TRACER improves model performance under rapidly evolving healthcare settings.

\section*{COMPETING INTERESTS}
None declared.

\section*{FUNDING}
This work was supported by the Duke/Duke-NUS Collaboration grant. The funder of the study had no role in the study design, data collection, data analysis, data interpretation, or writing of the report.

\section*{DATA AVAILABILITY STATEMENT}
Simulation code and data are available upon request. EHR data contain protected health information and are not able to be shared.

\section*{ACKNOWLEDGMENTS}
Health Data Science at Duke is supported by the National Center for Advancing Translational Sciences (NCATS), National Institutes of Health, through Grant Award Number UL1 TR002553. The Duke AI Health Data Science Fellowship Program is supported by the above grant, the Duke Department of Biostatistics \& Bioinformatics, and Duke AI Health. The Duke Protected Analytics Computing Environment (PACE) program is supported by the above grant and by Duke University Health System. The content of this publication is solely the responsibility of the authors and does not necessarily represent the official views of the NIH.

\printbibliography

\section*{Figure Legends}
\textbf{Figure 1:} Mixed data pattern caused by temporal shift due to COVID-19 pandemic. We only know when the pandemic (transition) starts, but don't know the transition status of each encounter.
\\
\textbf{Figure 2:} Proposed TRACER workflow. TRACER uses an EM algorithm to detect individual-level transitions and incorporates Trans-Lasso to adapt to evolving risk without retraining the full model.
\\
\textbf{Figure 3:} AUC comparisons across different models in the simulation study. Source/historical model: a model trained solely on historical data. Mixed/current model: a model trained solely on current pooled pre-transition and post-transition data.
\\
\textbf{Figure 4:} MSE comparisons across different models in the simulation study. /historical
model: a model trained solely on historical data. Mixed/current model: a model
trained solely on current pooled pre-transition and post-transition data.
\\
\textbf{Figure 5:} Comparison of predictive performance by month across different models in
the real data application. Monthly performance metrics (AUC, Brier score, and R²)
are shown for each model. Shaded bands represent 95\% bootstrap confidence intervals.
\\
\textbf{Figure 6:} Comparison of coefficients before and after transition in TRACER. Coef-
ficient estimates from the historical source model and post-transition model used in
TRACER are shown. Each point represents the effect size of a predictor. Differences
highlight changes in predictor importance across the transition.

\newpage
\appendix
\section{Assumptions in EM}
\label{sec:ass}
Outcome $Y_i$ is modeled as:

\begin{align}
P(y_i=1 | X_i, S_i=s)= \frac{\exp(\eta_s^{T}X_i)}{1+\exp(\eta_s^{T}X_i)},
\end{align}


where $\eta_0$ and $\eta_1$ are the parameters for outcome model before and after transition, respectively. In a transfer learning regime, we assume that the gap between the two latent models $\eta^1-\eta^0$, is close to $0$. To leverage this assumption, we could re-formulate the parameters as:
\[
\quad\eta_0,~\delta_{Y|X}=\eta_1-\eta_0.
\]
Then inspired by Trans-Lasso \cite{li_transfer_2022}, for high-dimensional features, we impose penalties on $\eta_0$, and $\delta_{Y|X}(t)$  by introducing $$\lambda_1\Psc_1(\eta_0)+\lambda_2\Psc_2(\delta_{Y|X}).$$ We could consider the sparse $\ell_1$ penalty on $\Psc_1(\eta_0)$ and $\Psc_2(\delta_{Y|X})$.

Meanwhile, we assume $P(S_{i}=1)=\pi(W_i)$ is determined by some covariates $W_i$ (such as age, gender, COVID diagnosis). We can specify the prior probability model for transition $\pi(W_i)$ as some parametric logistic model with parameter $\gamma$:
\begin{align}
    \pi(W_i)=\frac{\exp(\gamma^TW_i)}{1+\exp(\gamma^TW_i)},
\end{align}
and $\pi(W_i)=0$ if $t_i<T$.

For covariates, we suppose that there is a subset of covariates $A$ such that the distribution of which may differ from each other before and after transition. For example, we know that there will not be transition on features like age and gender for each subject. Density of $f(A_i|W_i,S_i=s)$ is assumed to be a Gaussian mixture, that is, $$A_i|W_i, S_i=0 \sim N(\mu_0, \Sigma_0)=N(\beta_0^T W_i, \Sigma_0)$$ and $$A_i|W_i,S_i=1 \sim N(\mu_1, \Sigma_1)=N(\beta_1^T W_i, \Sigma_1).$$

\newpage
\section{Computation - EM algorithm} 
\label{sec:EM}



Parameters $\gamma, \beta_0,\beta_1,\Sigma_0,\Sigma_1,\eta_0, \text{ and }\eta_1$ are needed to compute the posterior probability $P(S_i|W_i, A_i,Y_i=y)$ and the probability is used as the weight for TRACER. To estimate all the parameters, we can implement an EM-algorithm.

\paragraph{Initialization:}
First, $\hat\eta_0$ is obtained by regression on pre-transition historical data, and remain fixed during EM algorithm.
$$\hat\eta_0:Y_{i}^{\text{hist}}\sim Binomial(\frac{\exp(\eta_0^{T}X_i^{\text{hist}})}{1+\exp(\eta_0^{T}X_i^{\text{hist}})})$$

We initialize the parameters and transition $S$. Initial $\widehat S^{(0)}$ can be generated randomly from a binomial distribution, or in some cases, we might have a proxy of $S$ that can be used to initialize $\widehat{S}$. For example, we may use COVID diagnosis as the initial $S$.$\hat\gamma^{(0)}$ can be obtained by logistic regression $$\hat\gamma^{(0)}:\widehat S^{(0)}\sim \frac{\exp(\hat\gamma^{(0)T}W_i)}{1+\exp(\hat\gamma^{(0)T}W_i)}$$

Initialize $\hat\beta_0^{(0)}$ and $\hat\beta_1^{(0)}$ using linear regression 
$$\hat\beta_s^{(0)}:A\sim \hat\beta_s^{(0)T}W$$
for $s=0\text{ or }1$. Error term $\hat\epsilon_s=A_i-\hat\beta_s^{(0)T}W_i$ and $\widehat \Sigma_s^{(0)}$ is the covariance matrix of $\hat\epsilon_s$.

Then, iterate on E-step and M-step until converge or reach the the maximum number of iterations. For $t^{\text{th}}$ iteration:
\paragraph{E-step:} Given $\hat\gamma^{(t)}, \hat\beta_0^{(t)},\hat\beta_1^{(t)},\hat\Sigma_0^{(t)},\hat\Sigma_1^{(t)},\text{ and }\hat\eta_1^{(t)}$, impute the posterior probability of $S_i$ conditional on $W_i,A_{i}$ and $Y_i$.

When $Y_i=1$,
\begin{align*}
    &P(S_i|W_i, A_i,Y_i=1)\\
    &=\frac{P(S_i=1,A_i,Y_i|W_i)}{f(Y_i,A_i|W_i)}\\
    &=\frac{P(S_i=1,A_i,Y_i|W_i)}{f(Y_i,A_i|W_i,S_i=1)P(S_i=1|W_i)+f(Y_i,A_i|W_i,S_i=0)P(S_i=0|W_i)}\\
    &=P(S_i=1,A_i,Y_i|W_i)\\
    &/\{P(Y_i=1|A_i,W_i,S_i=1)f(A_i|W_i,S_i=1)P(S_i=1|W_i)\}\\
    &\{+P(Y_i=1|A_i,W_i,S_i=0)f(A_i|W_i,S_i=0)P(S_i=0|W_i)\}\\
    &=P(Y_i=1|A_i,W_i,S_i=1)f(A_i|W_i,S_i=1)P(S_i=1|W_i)\\
    &/\{P(Y_i=1|A_i,W_i,S_i=1)f(A_i|W_i,S_i=1)P(S_i=1|W_i)\}\\
    &\{+P(Y_i=1|A_i,W_i,S_i=0)f(A_i|W_i,S_i=0)P(S_i=0|W_i)\}\\
    &P_{S|W,A,Y=1}=\frac{P_{Y|A,W,S=1}f_{A|W,S=1}P_{S=1|W}}{P_{Y|A,W,S=1}f_{A|W,S=1}P_{S=1|W}+P_{Y|A,W,S=0}f_{A|W,S=0}P_{S=0|W}}
\end{align*}

where $$P(Y_i=1|A_i,W_i,S_i=s)=\frac{\exp(\eta_s^{(t)T}(A_i,W_i))}{1+\exp(\eta_s^{(t)T}(A_i,W_i))},$$

$$f(A_i|S_i=s,W_i)\sim N(\beta_s^{(t)T} W_i, \Sigma_s^{(t)}),$$

$$P(S_i=1|W_i)=\frac{\exp(\gamma^{(t)T}W_i)}{1+\exp(\gamma^{(t)T}W_i)},$$

and $$P(S_i=0|W_i)=1-P(S_i=1|W_i).$$

Similarly, when $Y_i=0$, $P(S_i|W_i, A_i,Y_i=0)$ can be estimated given those parameters.



\paragraph{M-step:} Update $\hat\gamma^{(t+1)}, \hat\beta_0^{(t+1)},\hat\beta_1^{(t+1)},\widehat\Sigma_0^{(t+1)},\widehat\Sigma_1^{(t+1)}, \text{ and }\hat\eta_1^{(t+1)}$. $\hat\eta_0$ is fixed given by the historical model.

Given $\hat p_s=P(S_i |W_i, A_i, Y_i=y_i)$ from E-step, solve $\hat\gamma^{(t+1)}$ using regression 
$$\hat\gamma^{(t+1)}:\hat p_s\sim \frac{\exp(\gamma^{(t+1)T}W_i)}{1+\exp(\gamma^{(t+1)T}W_i)}$$
 
Solve $\hat\beta_0^{(t+1)}$ and $\hat\beta_1^{(t+1)}$ using weighted linear regression
$$\hat\beta_s^{(t+1)}:A\sim \hat\beta_s^{(t+1)T}W$$
with weight $\hat p_s$ for $s=1$ and weight $1-\hat p_s$ for $s=0$. Error term $\hat\epsilon_s=A_i-\hat\beta_s^{(t+1)T}W_i$ and $\widehat \Sigma_s^{(0)}$ is the weighted covariance matrix of $\hat\epsilon_s$.

%

Then we estimate $\hat\eta_1^{(t+1)}=\hat\eta_0+\hat\delta_{Y|X}^{(t+1)}$ using \textbf{weighted Trans-Lasso}
(should not be too slow if offset not too correlated)



\begin{align*}
  \hat\delta_{Y|X}^{(t+1)}&\sim \frac{1}{N}\sum_{i=1}^N \hat p_s \ell(Y_{i},X_{i}^T\hat\eta_0)+(1-\hat p_s)\ell(Y_{i},X_{i}^T\{\hat\eta_0+\hat\delta_{Y|X}^{(t+1)}\}) + \lambda_2\Psc_2(\hat\delta_{Y|X}^{(t+1)}).  
\end{align*}

To increase the speed of EM algorithm when $A$ is large, we could reduce the dimension but preserve the relation between $A$ and $S$ by regress $S$ on $A$, and use $A_i^\prime=\widehat{P}(S_i|A_i)$ instead of $A_i$ when computing $f(A_i|S_i=s,W_i)$ and updating $\beta_s$ and $\Sigma_s$.

For a new prediction, where 
\begin{align}
&\widehat{P}(S_i=1|W_i,A_i)\\
&=\frac{f(A_i|W_i,S_i=1)P(S_i=1|W_i)}{f(A_i|W_i,S_i=1)P(S_i=1|W_i) +f(A_i|W_i,S_i=0)(1-P(S_i=0|W_i))}.
\end{align}

$$\widehat{P}(S_i=0|W_i)=\frac{\exp(\widehat\gamma^T\tilde W_i)}{1+\exp(\widehat\gamma^T\tilde W_i)},$$ and 
\begin{align*}
    f(A_i|W_i,S_i=s)=\phi_s(W_i)=\frac{1}{\sqrt{(2\pi)^n \, \det(\widehat{\Sigma}_s)}} 
  \exp\!\Bigl(-\tfrac{1}{2}(W_i-\hat{\mu}_s)^{T} \,\widehat{\Sigma}_s^{-1}\,(W_i-\hat{\mu}_s)\Bigr),
\end{align*}
where $\hat{\mu}_s=\hat{\beta}_s^TW_i$.

\newpage
\section{Predictors}
\label{predictors}

\begin{longtable}{ll}
\caption{Phecode and PhecodeString mapping for predictors used in the real data application.} \\
\hline
\textbf{Phecode} & \textbf{PhecodeString} \\
\hline
\endfirsthead

\multicolumn{2}{c}{{\bfseries Table \thetable\ (continued)}} \\
\hline
\textbf{Phecode} & \textbf{PhecodeString} \\
\hline
\endhead

\hline \multicolumn{2}{r}{{Continued on next page}} \\
\endfoot

\hline
\endlastfoot

41 & Bacterial infection NOS \\
465 & Acute upper respiratory infections of multiple or unspecified sites \\
480 & Pneumonia \\
250 & Diabetes mellitus \\
250.2 & Type 2 diabetes \\
250.22 & Type 2 diabetes with renal manifestations \\
261 & Vitamin deficiency \\
272 & Disorders of lipoid metabolism \\
275 & Disorders of mineral metabolism \\
276 & Disorders of fluid, electrolyte, and acid-base balance \\
276.1 & Electrolyte imbalance \\
278 & Overweight, obesity and other hyperalimentation \\
278.1 & Obesity \\
278.11 & Morbid obesity \\
279 & Disorders involving the immune mechanism \\
280 & Iron deficiency anemias \\
285.2 & Anemia of chronic disease \\
285.21 & Anemia in chronic kidney disease \\
286 & Coagulation defects \\
288 & Diseases of white blood cells \\
289 & Other diseases of blood and blood-forming organs \\
290 & Delirium dementia and amnestic and other cognitive disorders \\
296.2 & Depression \\
300 & Anxiety, phobic and dissociative disorders \\
306 & Other mental disorder \\
327 & Sleep disorders \\
327.3 & Sleep apnea \\
355 & Complex regional/central pain syndrome \\
338 & Pain \\
339 & Other headache syndromes \\
345 & Epilepsy, recurrent seizures, convulsions \\
386 & Vertiginous syndromes and other disorders of vestibular system \\
401.1 & Essential hypertension \\
401.22 & Hypertensive chronic kidney disease \\
401.2 & Hypertensive heart and/or renal disease \\
415 & Pulmonary heart disease \\
395 & Heart valve disorders \\
426 & Cardiac conduction disorders \\
427 & Cardiac dysrhythmias \\
427.2 & Atrial fibrillation and flutter \\
428 & Congestive heart failure; nonhypertensive \\
416 & Cardiomegaly \\
429 & Ill-defined descriptions and complications of heart disease \\
433 & Cerebrovascular disease \\
433.3 & Cerebral ischemia \\
452 & Other venous embolism and thrombosis \\
479 & Other upper respiratory disease \\
496.2 & Chronic bronchitis \\
495 & Asthma \\
510 & Other diseases of lung \\
509 & Respiratory failure, insufficiency, arrest \\
509.1 & Respiratory failure \\
512 & Other symptoms of respiratory system \\
573 & Other disorders of liver \\
578 & Gastrointestinal hemorrhage \\
585 & Renal failure \\
585.1 & Acute renal failure \\
585.3 & Chronic renal failure [CKD] \\
585.33 & Chronic Kidney Disease, Stage III \\
585.32 & End stage renal disease \\
588 & Disorders resulting from impaired renal function \\
591 & Urinary tract infection \\
599 & Other symptoms/disorders or the urinary system \\
619 & Noninflammatory female genital disorders \\
626 & Disorders of menstruation and other abnormal bleeding from female genital tract \\
707 & Chronic ulcer of skin \\
740 & Osteoarthrosis \\
741 & Symptoms and disorders of the joints \\
772 & Symptoms of the muscles \\
729 & Other disorders of soft tissues \\
771 & Musculoskeletal symptoms referable to limbs \\
771.1 & Swelling of limb \\
386.9 & Dizziness and giddiness (Light-headedness and vertigo) \\
783 & Fever of unknown origin \\
798 & Malaise and fatigue \\
350 & Abnormal movement \\
292 & Neurological disorders \\
782 & Symptoms involving skin and other integumentary tissue \\
687 & Symptoms affecting skin \\
427.7 & Tachycardia NOS \\
512.7 & Shortness of breath \\
512.9 & Other dyspnea \\
512.8 & Cough \\
418 & Nonspecific chest pain \\
790 & Nonspecific findings on examination of blood \\
790.6 & Other abnormal blood chemistry \\
585.31 & Renal dialysis \\
514 & Abnormal findings examination of lungs \\
401 & Hypertension \\
250.3 & Insulin pump user \\
457 & Encounter for long-term (current) use of anticoagulants, antithrombotics, aspirin \\
286.2 & Encounter for long-term (current) use of anticoagulants \\
457.3 & Encounter for long-term (current) use of aspirin \\
\end{longtable}

\newpage
\section{Full Coefficient List}
\label{coef_list}
\begin{figure}[htb]
    \centering
    \includegraphics[scale=1]{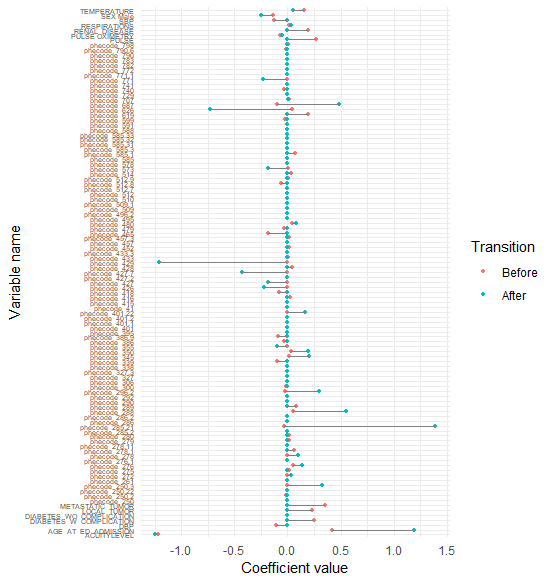}
    \caption{Coefficients of historical source model and post-transition model used in TRACER.}\label{coef_long}
\end{figure}

\end{document}